\documentclass[sigconf]{acmart}

\usepackage[utf8]{inputenc} 
\usepackage[T1]{fontenc}    
\usepackage{hyperref}       
\usepackage{url}            
\usepackage{booktabs}       
\usepackage{amsfonts}       
\usepackage{nicefrac}       
\usepackage{microtype}      
\usepackage{color}
\usepackage[normalem]{ulem}
\useunder{\uline}{\ul}{}
\usepackage{ulem}

\usepackage{graphicx}
\usepackage{subfigure}
\usepackage{booktabs} 
\usepackage{caption}
\usepackage{subfigure}

\usepackage{framed}

\usepackage{amssymb}
\usepackage{amsfonts}
\usepackage{mathrsfs}
\usepackage{mathtools}
\usepackage{array}
\usepackage{amsthm}
\usepackage{verbatim} 
\usepackage{enumerate}
\usepackage{bbm}
\usepackage{commath}
\usepackage{wrapfig}
\usepackage{amsbsy}
\usepackage{float}
\usepackage{amsmath}
\usepackage{algorithm}
\usepackage{tabularx}
\usepackage{listings}
\usepackage{xcolor}
\usepackage{enumitem}

\newcommand\blfootnote[1]{%
  \begingroup
  \renewcommand\thefootnote{}\footnote{#1}%
  \addtocounter{footnote}{-1}%
  \endgroup
}
\AtBeginDocument{%
  \providecommand\BibTeX{{%
    \normalfont B\kern-0.5em{\scshape i\kern-0.25em b}\kern-0.8em\TeX}}}

\copyrightyear{2022}
\acmYear{2022}
\setcopyright{acmcopyright}
\acmConference[WSDM '22] {Proceedings of the Fifteenth ACM International Conference on Web Search and Data Mining}{February 21--25, 2022}{Tempe, AZ, USA.}
\acmBooktitle{Proceedings of the Fifteenth ACM International Conference on Web Search and Data Mining (WSDM '22), February 21--25, 2022, Tempe, AZ, USA}
\acmPrice{15.00}
\acmISBN{978-1-4503-9132-0/22/02}
\acmDOI{10.1145/3488560.3498481}

\begin{document}
\fancyhead{}

\author[Z. Liu, Y. Luo, Z. Zelin, S. Li]{
    Zihan Liu$^{1,2}$, Yun Luo$^{2}$, Zelin Zang$^{2}$, Stan Z. Li$^{2\dag}$
}
\affiliation{
    $^1$ Zhejiang University \city{Hangzhou} \state{Zhejiang} \country{China}
}

\affiliation{
    $^2$ School of Engineering, Westlake University \city{Hangzhou} \state{Zhejiang} \country{China}
}

\email{
  {liuzihan,luoyun,zangzelin,stan.zq.li}@westlake.edu.cn
}

\fancyhead{}

\title{Surrogate Representation Learning with Isometric Mapping for Gray-box Graph Adversarial Attacks}

\begin{abstract}
Gray-box graph attacks aim to disrupt the victim model's performance by using inconspicuous attacks with limited knowledge of the victim model.
The details of the victim model and the labels of the test nodes are invisible to the attacker.
The attacker constructs an imaginary surrogate model trained under supervision to obtain the gradient on the node attributes or graph structure.
However, there is a lack of discussion on the training of surrogate models and the reliability of provided gradient information.
The general node classification models lose the topology of the nodes on the graph, which is, in fact, an exploitable prior for the attacker.
This paper investigates the effect of surrogate representation learning on the transferability of gray-box graph adversarial attacks.
We propose Surrogate Representation Learning with Isometric Mapping (SRLIM) to reserve the topology in the surrogate embedding.
By isometric mapping, our proposed SRLIM can constrain the topological structure of nodes from the input layer to the embedding space, that is, to maintain the similarity of nodes in the propagation process.
Experiments prove the effectiveness of our approach through the improvement in the performance of the adversarial attacks generated by the gradient-based attacker in untargeted poisoning gray-box scenarios.
\end{abstract}

\ccsdesc[300]{Computing methodologies~Supervised learning by classification}

\keywords{graph adversarial attack; gray-box attack; edge perturbation; representation learning; non-euclidean isometric mapping}

\maketitle

\section{Introduction}
Due to the advantages of modeling prior relationships between instances, graph-structured data is extensively applied in real-world domains such as social networks \cite{zhong2020multiple}, traffic networks \cite{wang2020traffic}, and recommendation systems \cite{wu2019session}.
Graph Neural Networks (GNNs), a deep learning network family based on graph-structured information aggregation \cite{zhou2020graph}, have empirically demonstrated excellent performance on downstream tasks such as node-level classification and graph-level classification. 
However, GNNs have proven to be vulnerable in the face of adversarial attacks in recent years \cite{dai2018adversarial,zugner2018adversarial}.
Even minor and imperceptible perturbations on graph data can mislead GNNs' predictions of target nodes or globally reduce the performance.
As a result, security and robustness on GNNs are speedily gaining research attention \cite{dai2018adversarial,zugner2018adversarial}. 

Generating perturbations to attack graph networks by adversarial learning has become an essential tool for analyzing the robustness and anti-interference ability of GNNs.
In the field of graph adversarial learning, researchers have proposed various settings for attacks.
According to the information available to the attackers, attacks can be divided into white-box (attackers can obtain all the information of the victim model) \cite{xu2019topology,wu2019adversarial}, gray-box (the parameters of the victim and the test labels are invisible) \cite{zugner2018adversarial,zugner2019adversarial,lin2020exploratory}, and black-box attacks (labels are invisible but attackers can do black-box queries to the prediction)  \cite{dai2018adversarial,ma2019attacking}, respectively.
The phase (i.e. model training or model testing) that an adversarial attack happens determines whether the attack is a poisoning attack or an evasion attack.

Among the proposed gray-box untargeted attack methods, the gradient-based attackers \cite{zugner2019adversarial,ma2020towards,xu2019topology,lin2020exploratory} have been proved to enhance attacking performance and become mainstream attack strategies.
A gradient-based attacker relies on a pre-trained GNN classifier, known as a surrogate model, to obtain gradients at node features and graph structure, based on which to generate perturbations.
However, existing papers have focused on using gradients to create perturbations and have not examined how to obtain more reliable gradients from the surrogate model.
We consider the representation learning process of the model as learning a mapping of nodes from the input layer to the embedding layer.
As a classifier trained with one-hot labels, GNN can map a node to the vicinity of the centroid of its corresponding cluster in the embedding layer, but it has difficulty capturing the topological relationships between nodes.
As shown in Figure \ref{topo}(a), the topology of nodes in the input layer is partially lost during the classification-based representation learning process, making the similarity between nodes in the input layer confusing when they are mapped in the embedding layer.\blfootnote{$^\dag$Corresponding author: Stan Z. Li.}
Based on the backpropagation of attack loss, the gradient on the input layer is directly related to the topology (similarity between nodes) in the embedding layer, while it is not entirely related to the topology in the input layer.
As a result, the gradient-based perturbations generated by the attacker are influenced by the mapping of the embedding layer of the surrogate model and become specific to this model while losing generalization to other models.
Considering the victim model is unknown, to improve the transferability of the attack, this paper proposes that the surrogate model should maintain the consistency of the topology of nodes in the embedding and input layers.
To this end, Surrogate Representation Learning with Isometric Mapping (SRLIM) has been proposed, which maintains the topology when mapping nodes from non-Euclidean input space to Euclidean embedding space.
We introduce the graph geodesic distance to represent the similarity between nodes in the input space and constrain the mapping of nodes to be isometric using an extra loss.
The surrogate model of SRLIM learns to maintain the node similarity from the input layer to the embedding layer. Thus the model establishes a connection between the derived gradient on the data level and the node topology on the input layer.

In short, the main contributions of this paper are summarized as follows: 
\vspace{-5pt}
\begin{itemize}[leftmargin=0.5cm]\setlength{\itemsep}{-1pt}
\item We discuss the shortcomings of the topology lost in representation learning when the surrogate model is a general GNN for gray-box attacks and propose that the surrogate model should retain the node topology from the input layer to the embedding layer.
\item We propose Surrogate Representation Learning with Isometric Mapping (SRLIM), enabling the model to learn topology.
It combines geodesic distances to estimate the similarity between nodes on the graph.
By maintaining the similarity of nodes from the input layer to the embedding layer, SRLIM can propagate topology information to the embedding layer.
\item We verify our approach's effectiveness by comparing our attack performance with those on other baseline models.
\end{itemize}

\section{Related Work}

There are three approaches to attack graph-structured data: modifying the node features \cite{ma2020towards,wu2019adversarial,zugner2018adversarial}, modifying the structure of the graph  \cite{dai2018adversarial,waniek2018hiding,zugner2019adversarial} and node injection \cite{sun2019node,sun2020adversarial}.
Each of them is bounded by a given budget, such as the matrix norm variation on features or the number of edge changes.
To analyze the robustness of GNNs, various attack methods under different settings have been proposed to explore the vulnerabilities of graph embedding methods and help develop the corresponding defense methods.
Attacks on the graph focus more on the main tasks such as classification \cite{dai2018adversarial,ma2020towards,xu2019topology}, community detection \cite{li2020adversarial,bojchevski2019adversarial}, and link prediction \cite{gupta2021adversarial,lin2020adversarial}.
Graph neural networks have been proven to be vulnerable and susceptible to attacks by Zugner et al. \cite{zugner2018adversarial}, who propose the earliest graph adversarial attack method Nettack for targeted node classification tasks.
In the same period, Dai et al. \cite{dai2018adversarial} propose a hierarchical Q Learning approach RL-S2V to reduce the overall node or graph classification accuracy by attacking the structure of graph data.
Subsequently, Wang et al. \cite{wang2018attack} propose a novel attack scenario in which adversarial artificial nodes and their edges are added without changing the original data to globally reduce the prediction accuracy of the network.
They use a greedy algorithm to generate links and corresponding attributes of artificial nodes and design a Generative Adversarial Networks based discriminator to ensure the imperceptibility of perturbations.
Some related articles \cite{sun2019node,sun2020adversarial} have also proposed to enhance node injection attack models based on reinforcement learning.

Zinger et al. \cite{zugner2019adversarial} propose Metattack, an attack model that modifies either the graph structure or node features by the meta-gradient of a trained surrogate model in the gray-box setup, which is the first gradient-based attack model.
Subsequently, papers \cite{xu2019topology,ma2020towards,lin2020exploratory,chang2020restricted} also take advantage of the gradient information provided by the surrogate in node features and graph structures to propose novel attack strategies and improved schemes.
These works focus on designing gradient-based gray-box attack strategies based on the meta-gradient obtained from a generally pre-trained surrogate model, a GNN. 
However, a neglected issue is that the GNN stucture, such as the Graph Convolutional Network (GCN) \cite{kipf2016semi}, GraphSage \cite{hamilton2017inductive}, and Chebnet \cite{defferrard2016convolutional}, is a factor affecting the gradient.
To be more precise, the structure (even the initialized weight parameters) of the GNN affects the representation learning process of the surrogate model, which leads to the different performance of the embedding layer.
As a result, when the network structure of the surrogate model is not the same as that of the unknown victim model, there will be a considerable variance in the embeddings of the networks, and the transferability of the attack will be limited.
This paper analyzes the impact of surrogate embedding on gradient-based attackers and illustrates the difficulties in gray-box attacks.

Inspired by visualization methods \cite{balasubramanian2002isomap,van2008visualizing,mcinnes2018umap}, our approach introduces isometric mappings to constrain the mapping of the network from the input layer to the embedding layer, allowing the network to retain topological information in forwarding propagation.

\section{Preliminaries and Definitions}
Before presenting the methodology and experiments, we first introduce preliminaries and definitions in this section.
Subsection 3.1 gives most of the notations used in the following sections.
Subsection 3.2 is the general background of GNN models and the node classification tasks. 
In Subsection 3.3, we present the method of deriving the meta-gradient of the adjacency matrix in a generic attack strategy with edge perturbations, which is initially proposed by Z\"{u}gner known as Metattack \cite{zugner2019adversarial}.

\subsection{Notations}
Given a graph with node attributes and a known subset of node labels, GNNs are expected to predict the class labels of unlabeled nodes. 
Let $G=(V,E,X)$ denotes an attribute graph, where $V=\{v_1,v_2,...,v_n\}$ is the set of $N$ nodes, $E \subseteq {V}\times{V} $ is the edge set and $X$ is the node feature matrix.
For the task of node classification, each node $i$ in the graph $G$ are associated with a feature matrix $x_i\in{\mathbb{R}^{d}}$ and a corresponding label $y_i\in{\mathbb{R}^{K}}$, where K is the number of classes.
We denote the topology structure of graph $G$ as a binary adjacent matrix $A = \{0,1\}^{{N}\times{N}}$, where $A_{i,j}=1$ if and only if $(i,j)\in{E}$.
Moreover, we denotes the learning models using $f_\theta (\cdot)$ with parameters $\theta$. 
We denote the representation in the hidden layer with $H$ and the weight matrix of a linear transformation with $W$.
In addition, we use $Z$ to represent the predicted probability of $f_\theta$ and $\mathcal{L}$ to represent the loss function.
We use  $'$  to distinguish it from the original, written as $G'=(A',X')$ for a perturbed graph.
The attacker's perturbation budget on a graph is denoted as $\Delta$.

\subsection{GNN Models}
For the graph $G$, a GNN learns a function $f_{\theta}(G)$ that can map a node features into a possibility distribution $Z=\{z_1,z_2,...z_K\}$ of $K$ classes.
Formally, the $l$-th layer of a GNN model consisted of $L$-layer follows the general form:
\begin{equation}
    H^{(l)}=\sigma{(\hat{A}H^{(l-1)}W^{(l-1)})},
\end{equation}
where $H^{(l)}\in{\mathbb{R}^{{N}\times{D_l}}}$ denotes the latent representation at layer $l$ ($H^{(0)}$ refers to the input features), $W_{(l-1)}\in{\mathbb{R}^{{d^{(l-1)}}\times{d^{(l)}}}}$ is the weight matrix at $(l$-$1)$-th layer and $\sigma$ is a nonlinear activation function.
$\hat{A}=\tilde{D}^{\frac{1}{2}}\tilde{A}\tilde{D}^{\frac{1}{2}}$ is the normalized adjacent matrix, where $\tilde{A}=A+I_N$ is the adjacent matrix with self-loops , and $\tilde{D}=\sum_j{\tilde{A}_{i,j}}$ represents the degree matrix of $G$.
Taking a two-layer GCN as an example, the network output for a node classification task is expressed as follows:
\begin{equation}
    Z=f_{\theta}(A,X)=softmax(\hat{A}\sigma{(\hat{A}XW^{(0)})}W^{(1)}),
\end{equation}
The parameters $\theta$ of function $f_\theta$ are iteratively trained by minimizing a cross-entropy loss $\mathcal{L}_{ce}$ on the nodes in training set:
\begin{equation}\label{3}
{\theta}^*=\underset{\theta}{\mathrm{argmin}}\, \mathcal{L}_{train}(f_\theta(G))=
\underset{\theta}{\mathrm{argmin}}
\sum_{i=1}^{N}\mathcal{L}_{ce}(z_i,y_i).
\end{equation}
The GNN model $f_{{\theta}^*}$ with parameter ${\theta}^*$ is a trained classifier for node classification task.
\subsection{Gradient-based edge perturbations}
In the scenario where the attacker performs edge perturbations on the graph, the perturbations occur only on the adjacent matrix $A'$.
The $l_0$ norm of the variation of the perturbation graph relative to the original graph is constrained by the perturbation budget $\Delta$, written as:
\begin{equation}
\lVert A'-A\rVert_0 \leq 2\Delta,
\end{equation}
where the budget $\Delta$ less than $5\%$ is considered to be inconspicuous in common.
We consider the graph under the scenario to be an undirected graph, so the upper limit is $2\Delta$.
For untargeted attacks, the attacker aims to insert poisons to full the victim model by globally reducing the overall performance on test nodes on the perturbed graph $G'$.

Meta-gradients on graph structures are widely used for gray-box attackers \cite{zugner2019adversarial,xu2019topology,lin2020exploratory}.
Since the victim model is agnostic, attackers require training a surrogate model, generally a member of the GNN family, to obtain the gradient information at the input layer through back-propagation.
The surrogate model requires proper pre-training in order to simulate the training process of the victim model.
Unlike the gradient-based attack methods that are widely used in the field of computer vision \cite{yuan2019adversarial}, for graph-structured data, the discretization of the graph structure dictates that the gradients on the structure cannot be directly applied to the adjacency matrix.
As a compromise, the gradient on the adjacent matrix is considered a reference book to help find edge flipping candidates that are likely to be ideal perturbations.
The following formula can calculate the gradient on the graph structure:
\begin{equation}
    A^{grad}=\nabla_{A}\mathcal{L}_{atk}(f_{\theta^*}(G)),
\end{equation}
where $\mathcal{L}_{atk}$ is the attacking loss function and $f_{\theta^*}$ is the surrogate model trained by the objective function in equation \ref{3}.
To lower the classification accuracy, $\mathcal{L}_{atk}$ has two forms.
One is $\mathcal{L}_{atk}=-\mathcal{L}_{train}$ for reducing the predicted probability of labeled classes, where the loss is calculated by few labeled nodes.
The other one is $\mathcal{L}_{atk}=-\mathcal{L}_{self}$, which is applied to all nodes to reduce the confidence on the original prediction using pseudo-labels.
For the edge between nodes $v_i$ and $v_j$, if $A_{i,j}=1$ and $A_{\,i,j}^{grad}<0$, or if $A_{i,j}=0$ and $A_{\,i,j}^{grad}>0$, then flipping the edge $E_{i,j}$ will be more likely to degrade the overall performance of the victim model.
For an attack model $\varphi$, its process of generating an perturbed graph $G'$ from gradient information can be expressed as:
\begin{equation}\label{grad_pert}
    G'=\varphi(G,A^{grad})=\varphi(G,\nabla_{A}\mathcal{L}_{atk}(f_{\theta^*}(G))).
\end{equation}
It can be seen from equation \ref{grad_pert} that, as the only reference of the attack model, gradient directly affects the performance of the attacker.
Therefore, inspired by the impact of the surrogate node embedding on the generated adversarial sample, this paper aims to optimize the surrogate model to provide a stronger structural gradient for the attack strategy to improve the attack success rate.
In the following sections, we will discuss the adverse effects on the attacker and gradient caused by the drop of topological information during the learning process of the surrogate model and the methods we propose to solve this problem.

\section{Methodology}

In this section, we begin by discussing the impact of the drop of topological information in the input layer on the gradient as well as on the transferability of the attack.
We then demonstrate our approach for enabling the node features of the embedding layer to preserve the topology in the input layer during the training of the surrogate model.

\subsection{Why Topology Matters?}

The surrogate model $f_\theta$ is trained by the downstream node classification task, where the loss function is cross-entropy, and the labels are in the form of one-hot.
The training of the surrogate model follows Eq (\ref{3}) and outputs model $f_{\theta ^*}$ which is used to derive the gradients on the input.
Assume that there are nodes $v_a$, $v_b$, $v_c$ and clusters $c_\alpha$,$c_\beta$,$c_\gamma$. Each cluster represents the distribution of nodes of a category at the input or embedding space.
Nodes $v_a$, $v_b$ and $v_c$ belongs to the cluster $c_\alpha$, $c_\beta$ and $c_\gamma$, respectively.

   \begin{figure}[thpb]
      \centering
      \includegraphics[width =  \hsize]{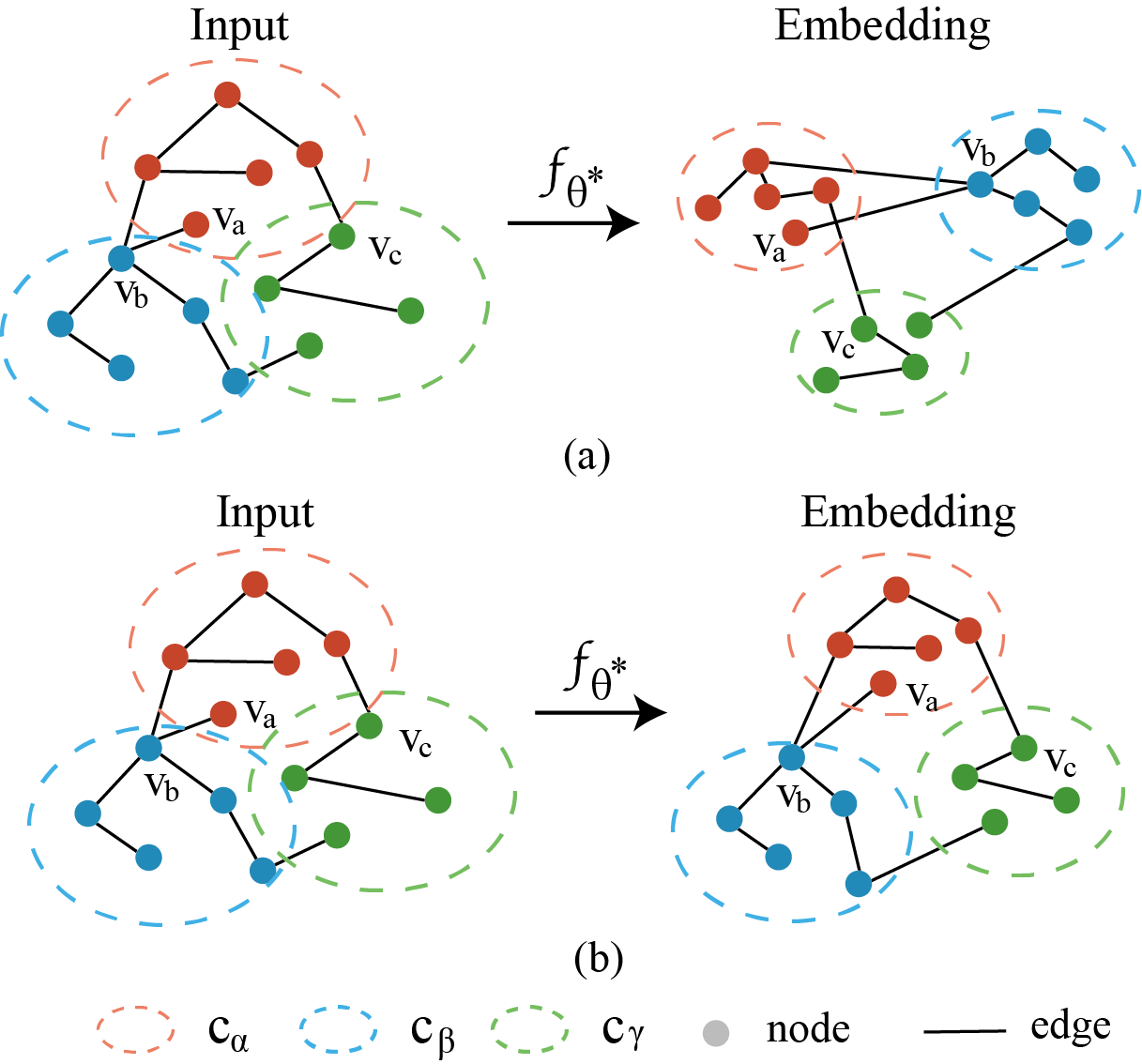}
      \caption{Schematic diagram of the surrogate model mapping node from the input layer to the embedding layer. In case (a), the topology is lost during the mapping process. In case (b), topology information is retained during the mapping process.}
      \label{topo}
   \end{figure}

Figure \ref{topo} shows the schematic diagram of the mapping of $f_{\theta ^*}$ from the input layer to the embedding layer.
The figure contains two cases (a) and (b), corresponding to whether the model preserves the topological information of the nodes in the mapping process.
During the training process, the model learns to map the nodes with scattered distribution in the input space to the clusters of the corresponding classes on the embedding layer. 
From the view of node distribution at the input layer, misleading the model to classify node $v_a$ toward the direction of $v_b$ (or into the area of $c_\beta$) will lead to a lower cost.
In the mapping process of Figure 1 (a), node $v_a$ is mapped to a position close to $v_c$ and $c_\gamma$ in the  embedding layer.
The attack model tends to misdirect $v_a$ towards $c_\gamma$ because $v_a$ and $c_\gamma$ are closer in the embedding space thus the gradient in that direction is more significant.
It is worth noting that the model's predictions are directly related to the embedding layer of the model, both in terms of the forward process started from the input and backpropagation started from the output.
When the network loses topological information during the forward process, the gradients derived from the attack loss by backpropagation also do not contain the topological information of the input layer.
However, the topology of the nodes in the input space is a priori valuable to the attacker.
Considering that the victim model is unknown, the network should keep the topology consistent during the propagation to improve the generalization and transferability of the attack, as shown in Figure \ref{topo}(b).
Learning the topology of nodes simultaneously during representation learning makes the gradient of the attack loss correlate to the topology of nodes on the input layer.
In this way, the gradient from the surrogate model contains a priori knowledge of the topological relationships of the nodes in the graph, thus making it more reliable to attackers.

\subsection{Non-Euclidean Similarity Estimation}
The graph structure data is distributed in a non-Euclidean space because nodes have adjacency relationships with other nodes in addition to their attributes.
Traditional similarity metrics, such as Euclidean Distance and Manhattan Distance, do not characterize the graph's topology.
To obtain an estimated representation of the similarity between nodes in the non-Euclidean space, we introduce the connectivity and geodesic distance of the nodes in the graph.

We start by considering the connectivity between the nodes.
For nodes that are not connected in the graph, the distance before them is infinity.
For nodes connected in the graph, the distance between them is the sum of the lengths of the edges that connect them in the shortest path. 
The shortest path between node $v_i$ and node $v_j$ consists of $k$ edges, denoted as $\Phi(i,j)=\{E_{(\phi_0,\phi_1)},E_{(\phi_1,\phi_2)},...,E_{(\phi_{k-1},\phi_k)}\}$, where $E$ stands for the edge between two nodes, $\phi_0$ corresponds to node $v_i$ and $\phi_k$ corresponds to node $v_j$.
The geodesic distanced from $v_i$ to $v_j$, $\tau{(v_i,v_j)}$, is calculated by the following formula:

\begin{equation}
\tau{(v_i,v_j)}=\left\{
\begin{aligned}
&\sum\limits_{p=0}^{k-1} d(E_{(\phi_p,\phi_{p+1})})     &&& {\rm if} \; A^\infty_{\;i,j}>0 \\
&\gamma\; max([\tau{(v_i,v_j)}])     &&& {\rm if} \; A^\infty_{\;i,j}=0
\end{aligned}
\right.
\end{equation}

where $d(E)$ represents the cosine similarity of the two nodes directly connected by the edge $E$ on the path and $A^\infty_{\;i,j}>0$ represents $v_i$ and $v_j$ are connected on the graph.
The formula means, if nodes $v_i$ and $v_j$ are connected on the graph, the distance between them is the accumulated distance of the edges in $\Phi(i,j)$.
The maximum distance between connected nodes is multiplied by an enormous constant term $\gamma$ as the distance between all disconnected nodes on the graph.

In order to avoid the adverse effects of outliers and neighborhood inhomogeneity in the real-world data, refering to \cite{mcinnes2018umap}, the geodesic distance $\tau{(v_i,v_j)}$ is transformed by the following equations:

\begin{equation} \label{T}
    T(\tau,u)=\sqrt{2\pi}\cdot \frac{\Gamma(\frac{u+1}{2})}{\sqrt{u\pi}\cdot \Gamma(\frac{u}{2})} \Big( 1+\frac{\tau^2}{u} \Big)^{-\frac{u+1}{2}}
\end{equation}
\begin{equation} \label{epsilon}
    \epsilon^*_i=\underset{\epsilon_i}{\mathrm{argmin}}\; \Big| Q-2^{\sum_j T(\frac{\tau{(v_i,v_j)}-\xi_i}{\epsilon_i},u)^2} \Big|
\end{equation}
\begin{equation} \label{distance}
    \tilde\tau{(v_i|v_j)} = \frac{\tau{(v_i,v_j)}-\xi_i}{\epsilon^*_i},
\end{equation}
where in Eq. \eqref{T} $T$ is a $t$-distribution density function mapping the geodesic distance to similarity, $\Gamma$ is the gamma function and $u$ is the degree of freedom in $t$-distribution; in Eq. \eqref{epsilon} and Eq. \eqref{distance}, $\epsilon^*_i$  is the optimal solution of the objective function determined by the binary search method, the objective function controls the compactness of $v_i$ and its neighbors by hyper-parameters $Q$ and $u$, $\xi_i=min([\, \tau(v_i,v_j)\mid j=1,...,n\,])$ is the minimum term deducted from the distance between $v_i$ and other nodes in order to mitigate possible skewed embedding caused by outliers and $\tilde\tau{(v_i|v_j)}$ is the transformed distance from $v_i$ to $v_j$.
It is worth noting that $\tilde\tau{(v_i|v_j)}$ is unequal to $\tilde\tau{(v_j|v_i)}$ since they have different $\xi$ and $\epsilon^*$.

Further, the symmetric geodesic similarity between $v_i$ and $v_j$ can be calculated as follows:
\begin{equation} \label{s1}
    s_{v_i|v_j} = T(\tilde\tau{(v_i|v_j)},u) 
\end{equation}
\begin{equation} \label{s2}
    s_{i,j} = s_{v_i|v_j}+s_{v_j|v_i}-2\,s_{v_i|v_j}\,s_{v_j|v_i}.
\end{equation}
Thus, we can obtain a graph geodesic similarity matrix $S(A,X^*)=\{s_{i,j}\mid i,j=1,...,n\}$, where $X^*$ can be either node attributes $X$ or representations in any latent layer $H^{(l)}$.

   \begin{figure*}[thpb]
      \centering
      \includegraphics[scale=1.0]{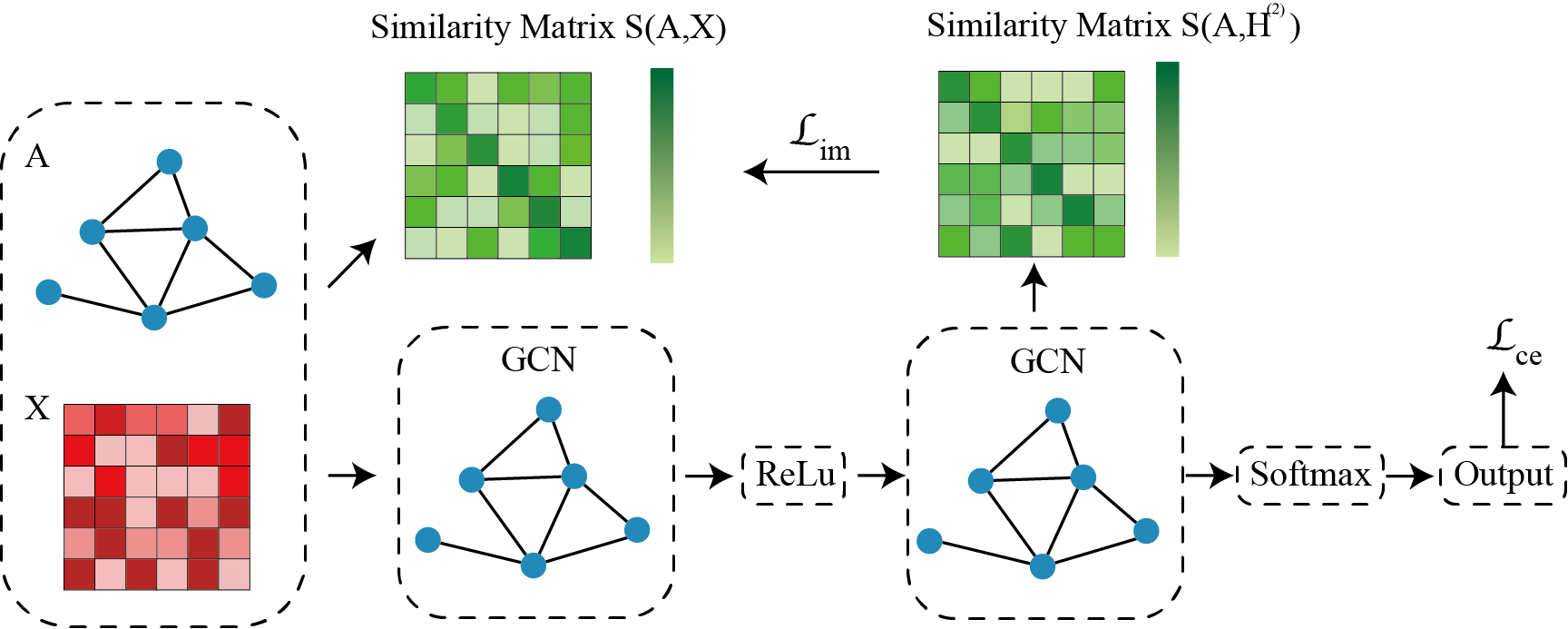}
      \caption{An Implementation on GCN Framework of the proposed SRLIM.}
      \label{model}
   \end{figure*}

\subsection{Isometric Mapping}
By calculating the graph geodesic similarity matrix $S(A,X^*)$, we can derive the topology of nodes at any layer of the surrogate model.
Inspired by \cite{balasubramanian2002isomap}, we use isometric mapping to propagate the topology between layers of a deep network.
Specifically, we use the geodesic similarity matrix $S(A,X)$ on the input layer to constrain $S(A,H)$ in the hidden layer.

Figure \ref{model} illustrates the overall framework for training the surrogate model using SRLIM.
The network structure of the surrogate model consists of a two-layer GCN.
The representations of the nodes on the embedding layer output the predicted probability via a Softmax function.
In the training process, we introduce an isometric mapping loss $\mathcal{L}_{im}$ to fit these two similarity matrices in addition to the cross-entropy loss. 
The loss function of the surrogate model based on SRLIM is expressed as:
\begin{equation}
    \mathcal{L}_{SRLIM}=\mathcal{L}_{ce}(Z,Y)+\lambda \, \mathcal{L}_{im}(S(A,X),S(A,H^{(L)})),
\end{equation}
where $Z$ is the probability distribution, $Y$ is the set of labels, and $H^{(L)}$ is the last hidden layer before the output of the network, which is the node embedding.
The first term, $\mathcal{L}_{ce}$, is the cross-entropy function commonly used in the training phase of supervised node classification tasks.
The second term $\mathcal{L}_{im}$ is the Bergman divergence of the similarity of the nodes in the input space and the embedding space.
The Bergman divergence is defined as follows:
\begin{equation}
    \mathcal{L}_{im}(\, \alpha,\beta\mid F(\cdot)) = F(\alpha) - F(\beta) - \langle \bigtriangledown F(\alpha),\alpha-\beta \rangle,
\end{equation}
where $F(\cdot)$ is a continuously differentiable convex function defined on a closed convex set.
The form of Bergman divergence used in SRLIM is defined as:
\begin{equation}
\begin{split}
    & \mathcal{L}(\, \alpha,\beta\mid \alpha\,\log (\alpha)+(1-\alpha)\log (1-\alpha)) = \\
    & \qquad \quad \alpha\, \log \frac{\alpha}{\beta}+(1-\alpha)\log \frac{1-\alpha}{1-\beta} \;.
\end{split}
\end{equation}
The inputs of $\mathcal{L}_{im}$ are the similarity matrix of data features $S(A,X)$ and embedding $S(A,H^{(L)})$.
The purpose of the second term in $\mathcal{L}_{SRLIM}$ is to reduce the difference in the topology of nodes between the two layers. 
Since $S(A,X)$ is unlearnable, the surrogate model learns to fit the embedding topology $S(A,H^{(L)})$ to that of the input.
When the second term converges, we consider that the isometric mapping happens in the surrogate model, that is, the surrogate model can learn the topology from the original data.

Considering that the computational complexity of the second term is $O(N^2)$, we use local isometric mapping instead of global isometric mapping.
We extract nodes from the dataset in batches and compute the similarity between small-scale samples. 
The computational complexity required to iterate through all nodes is 
$O(\frac{N}{bs}\cdot {bs}^2)=O(N\cdot bs)$ ,where $bs$ represents the batch size.
To ensure that local isometry extends globally, we rearrange the combination of nodes in batches after each epoch.

\section{Experiments}
\label{sec_exp}

\begin{table}[]
\caption{Statistics of datasets.}
\label{stat}
\begin{tabular}{lcccc}
\hline
Datasets & \multicolumn{1}{l}{Vertices} & \multicolumn{1}{l}{Edges} & \multicolumn{1}{l}{Classes} & \multicolumn{1}{l}{Features} \\ \hline
Citeseer & 3312                         & 4732                      & 6                           & 3703                         \\
Cora     & 2708                         & 5429                      & 7                           & 1433                         \\
Cora-ML  & 2995                         & 8416                      & 7                           & 2879                         \\ \hline
\end{tabular}
\end{table}

\begin{table*}[]
\begin{center}
\caption{Attack results under gray-box setup when the surrogate and victim models are consisted of GCN. The group 'Original' denotes the result for unperturbed data. The table shows the misclassification rate (\%) for SRLIM and other baselines under perturbation rates 1\%, 3\% and 5\% on Citeseer, Cora and Cora-ML. The best results from each experiment are bold.}
\label{table1}
\setlength{\tabcolsep}{3.2mm}{
\begin{tabular}{cccccccccc}
\hline
{\ul \textit{}}                 & \multicolumn{3}{c}{Citeseer}                                                  & \multicolumn{3}{c}{Cora}                                                      & \multicolumn{3}{c}{Cora-ML}                                                   \\ \hline
\multicolumn{1}{l}{Pert Rate} & \multicolumn{1}{l}{1\%} & \multicolumn{1}{l}{3\%} & \multicolumn{1}{l}{5\%} & \multicolumn{1}{l}{1\%} & \multicolumn{1}{l}{3\%} & \multicolumn{1}{l}{5\%} & \multicolumn{1}{l}{1\%} & \multicolumn{1}{l}{3\%} & \multicolumn{1}{l}{5\%} \\ \hline
Original                           & \multicolumn{3}{c}{32.2}                                                      & \multicolumn{3}{c}{18.9}                                                      & \multicolumn{3}{c}{17.7}                                                      \\ \hline
DICE                            & 32.2                     & 32.4                     & 32.7                     & 19.1                     & 19.4                     & 20.1                     & 17.8                     & 18.2                     & 19.0                     \\
Meta-Train                      & 34.2                     & 35.1                     & 36.7                     & 21.8                     & 22.3                     & 24.2                     & 17.9                     & 18.9                     & 23.1                     \\
Meta-Self                       & 33.6                     & 35.0                     & 38.5                     & 20.3                     & 23.4                     & 26.8                     & 18.7                     & 19.4                     & 21.9                     \\
EpoAtk                          & 35.8                     & 37.3                     & 39.4                     & 22.3                     & 25.7                     & 29.6                     & 20.3                     & 25.5                     & 28.4                     \\ \hline
SRLIM                           & \textbf{36.1}            & \textbf{38.7}            & \textbf{40.1}            & \textbf{22.9}            & \textbf{26.3}            & \textbf{30.8}            & \textbf{20.9}            & \textbf{28.0}            & \textbf{29.4}            \\ \hline
\end{tabular}
}
\end{center}
\end{table*}

\begin{table*}[]
\begin{center}
\caption{Attack results under gray-box setup when the surrogate model is consisted of GCN and the victim is consisted of ChebNet.}
\label{table2}
\setlength{\tabcolsep}{3.2mm}{
\begin{tabular}{cccccccccc}
\hline
          & \multicolumn{3}{c}{Citeseer} & \multicolumn{3}{c}{Cora} & \multicolumn{3}{c}{Cora-ML} \\ \hline
Pert Rate & 1\%      & 3\%      & 5\%     & 1\%     & 3\%    & 5\%    & 1\%      & 3\%     & 5\%     \\ \hline
Original  & \multicolumn{3}{c}{31.5}     & \multicolumn{3}{c}{20.6} & \multicolumn{3}{c}{18.2}    \\ \hline
Meta-Self & 31.7     & 32.7     & 33.0    & 20.9    & \textbf{22.7}   & 24.3   & 18.6     & \textbf{19.9}    & \textbf{21.2}    \\
EpoAtk    & 31.3     & 32.5     & 33.3    & 21.2    & 22.3   & 24.0   & \textbf{18.8}     & 19.1    & 19.5    \\ \hline
SRLIM     & \textbf{32.0}     & \textbf{33.2}     & \textbf{33.9}    & \textbf{21.4}    & 22.1   & \textbf{25.1}   &  \textbf{18.8}        & 19.3        & 19.6         \\ \hline
\end{tabular}
}
\end{center}
\end{table*}

This section has designed experiments to evaluate SRLIM by comparing it with the state-of-the-art baselines on well-known datasets.
Each attacker consists of a surrogate model and an attack model. 
To test our approach, we combine our proposed surrogate model with the attack model from \cite{lin2020exploratory}, which forms the attacker SRLIM.
We begin by describing the datasets, the baselines, and the experimental setup we used in our experiments. 
Then we aim to discuss the performance of SRLIM in a scenario under a gray-box attack and attempt to demonstrate its effectiveness.

\subsection{Datasets}
We adopt three citation network datasets: Citeseer \cite{sen2008collective}, Cora \cite{mccallum2000automating} and Cora-ML \cite{mccallum2000automating} to evaluate our approach.
For these graphical datasets, vertices represent papers, and edges represent citations.
In addition, each node is described by a bag-of-words feature. 
Their specific details are listed in Table \ref{stat}.
Following the experimental setup in \cite{zugner2019adversarial}, all datasets are randomly split to 10\% of labeled nodes and 90\% of unlabeled nodes.
The labels of unlabeled nodes are invisible to both the attacker and the surrogate and are only available when evaluating the performance of the adversarial attacks.

\subsection{Baselines}
DICE \cite{waniek2018hiding}, Meta-Train\cite{zugner2019adversarial}, Meta-Self \cite{zugner2019adversarial}, EpoAtk\cite{lin2020exploratory} are baselines, all of which attack GNNs by edge perturbations.
The implementation of baselines is detailed as follows.

\textbf{DICE} A method based on randomness, which removes the edges between nodes from the same classes and adds the edges between nodes from different classes. For an unlabeled node, the highest confidence of its prediction is considered as its label.

\textbf{Meta-Self \& Meta-Train} Gradient-based attackers that treat the graph structure as a hyper-parameter to optimize via meta-learning. Two attack models differ in their attack loss function.

\textbf{EpoAtk} An enhanced meta-gradient based method. EpoAtk's exploration strategy consists of three phases: generation, evaluation, and recombination, aiming to avoid the possible error incorrect messages from the gradient information.

\subsection{Experimental Settings}
We conduct the experiments in the untargeted poisoning gray-box setup.
In terms of gray-box, the setup is further split into two scenarios.
The gray box indicates the degree of openness to information about the victim model.
We discuss both network architecture agnostic and weight parameter agnostic scenarios for the victim model.
In the scenario of weight parameter agnostic, attackers know the components of the network and the number of neurons in each layer.
The implementation of this testing scenario follows \cite{lin2020exploratory} \footnote{This scenario is strictly between white-box attack and gray-box attack because the attacker knows the details of the victim model (such as the width of the linear layer and the type of the activation function), which should be unknown in a gray-box attack.}.
Because the initial value of the weight parameter and other training details are unknown, it is yet impossible to simulate the victim model in such scenarios.
In the scenario of network architecture agnostic, there is no basis for the attacker to construct a surrogate model.

The experiment compares the attack performance of each method under the perturbation rate of 1\%, 3\%, and 5\%.
We repeat the same experiment 20 times with specific seeds for each result shown in the tables below, where the maximum and minimum values have been removed.
The results show the mean value of the misclassification rate of node classification tasks.
To ensure a fair comparison, the network composition, training duration, and optimization of the victim networks are identical for different experiments under the same experimental setup.

\subsection{Attack Performance}

We have designed experiments to compare SRLIM with some baselines and present the results in this section.
We conduct separate experiments in both the network architecture agnostic scenarios and the weight parameter agnostic scenarios.
Both experimental scenarios are designed to verify the transferability of the attack method.
When the network composition of the surrogate model and the victim model is the same, the performance of their embedding layer will have some similarities.
In this scenario, the difficulty of migration attack is relatively simple compared to the other scenario.
The attacking budget $\Delta$ is set at 1\%, 3\%, and 5\% of edges in the original graph in the experiments.
The control group 'Original' represents the misclassification rate of the model trained using an unperturbed graph under the same training process.

\subsubsection{Weight Parameter Agnostic}

The results of the attack when the network architecture of the victim model is available to the surrogate model are shown in Table \ref{table1}.
Both the surrogate and the victim models are set as a GCN but with different initialization of weight parameters.
In this scenario, the attacker obtains more information about the victim model, so the transferability of the attack will be stronger.
From table \ref{table1} we can observe that SRLIM outperforms the existing state-of-the-art baselines in all untargeted poisoning attack experiments.
From the perspective of the overall attack performance, the effect of SRLIM is the most significant, followed by EpoAtk, the results of Meta-Self and Meta-Train are close, while DICE is the worst among all baselines.
Our method has improvements over other methods, especially on the Cora-ML, with enhancements of 0.6\%, 2.5\%, and 1.0\% under the perturbation rates of 1\%, 3\%, and 5\%, respectively.
On the Cora, the result of SRLIM outperforms the second-best method by 0.6\% to 1.2\%.
On Citeseer, SRLIM performs better than the second-best method by 0.3\% to 1.4\%.
From the results shown in Table \ref{table1}, the performance of SRLIM exceeds that of other state-of-the-art methods.
The results demonstrate that the training of the surrogate model does influence the attacker's performance. 
The results also show that our proposed SRLIM, a representation learning method for non-Euclidean space isometric mapping, provides more robust and reliable gradient information when the model architecture is visible to the attacker.

\subsubsection{Network Architecture Agnostic}

Table \ref{table2} shows the experimental results in the scenario where the victim model is agnostic to the attacker.
The surrogate model is consisted of GCN, while the victim model is set as ChebNet.
Results show the performance of Meta-Self, EpoAtk, and RALoG at different perturbation rates on Citeseer, Cora, and Cora-ML, respectively.
Comparing the results in Table \ref{table2} with that in Table \ref{table1}, although both scenarios belong to gray-box setup, all the attackers perform much worse when the architecture of the victim model is unknown. 
It also proves that the differences in the performance of the embedding layers of different network architectures are significant, and this is one of the critical factors that hinder the transferability of the attack.
Our aim in proposing SRLIM is to improve the generalization of the gradients obtained from the surrogate model so that gradient-based attacks have a greater possibility of being effective against other models.

The results show that SRLIM is more effective than other attack methods in most experiments in Table \ref{table2}.
On the Citeseer, SRLIM outperforms the second-place attacker 0.3\%, 0.5\% and 0.6\% at perturbation rates of 1\%, 3\%, and 5\%, respectively.
On the Cora dataset, SRLIM surpasses the second place 1\% and 5\% with perturbation rates of 0.2\% and 0.8\%.
Meta-Self performs better than others at a 3\% perturbation rate on Cora with a misclassification rate of 22.7\%.
On the Cora-ML, SRLIM and EpoAtk perform best at a 1\% perturbation rate with misclassification rates of 18.8\%, and Meta-Self outperforms the others at 19.9\% and 21.2\% with perturbation rates of 3\% and 5\%.
Regarding the performance of Meta-Self on Cora-ML due to our model, We believe the reason is that SRLIM borrows the attack module from EpoAtk. 
From the experimental results, we can see that EpoAtk does not perform as well as Meta-Self on Cora-ML. 
Compared to EpoAtk, SRLIM has demonstrated that our proposed surrogate model improves the overall attack transferability.
In addition, it is worth noting that when we measure the effectiveness of an attack, we need to subtract the misclassification rate of the original model from that of the perturbed. 
In the results shown in Table \ref{table2}, even a 0.1\% improvement is significant.
Therefore, the enhancement of SRLIM is notable compared to other attack methods.

Through the experiments shown in Sections 5.4.1 and 5.4.2, we demonstrate that our proposed SRLIM, which learns the topology utilizing isometric mapping, can improve the reliability and robustness of the gradient information and thus the transferability of the poison attack under the gray box setting.

\subsection{Visualization of Embeddings}

   \begin{figure}[thpb]
      \centering
      \includegraphics[scale=0.65]{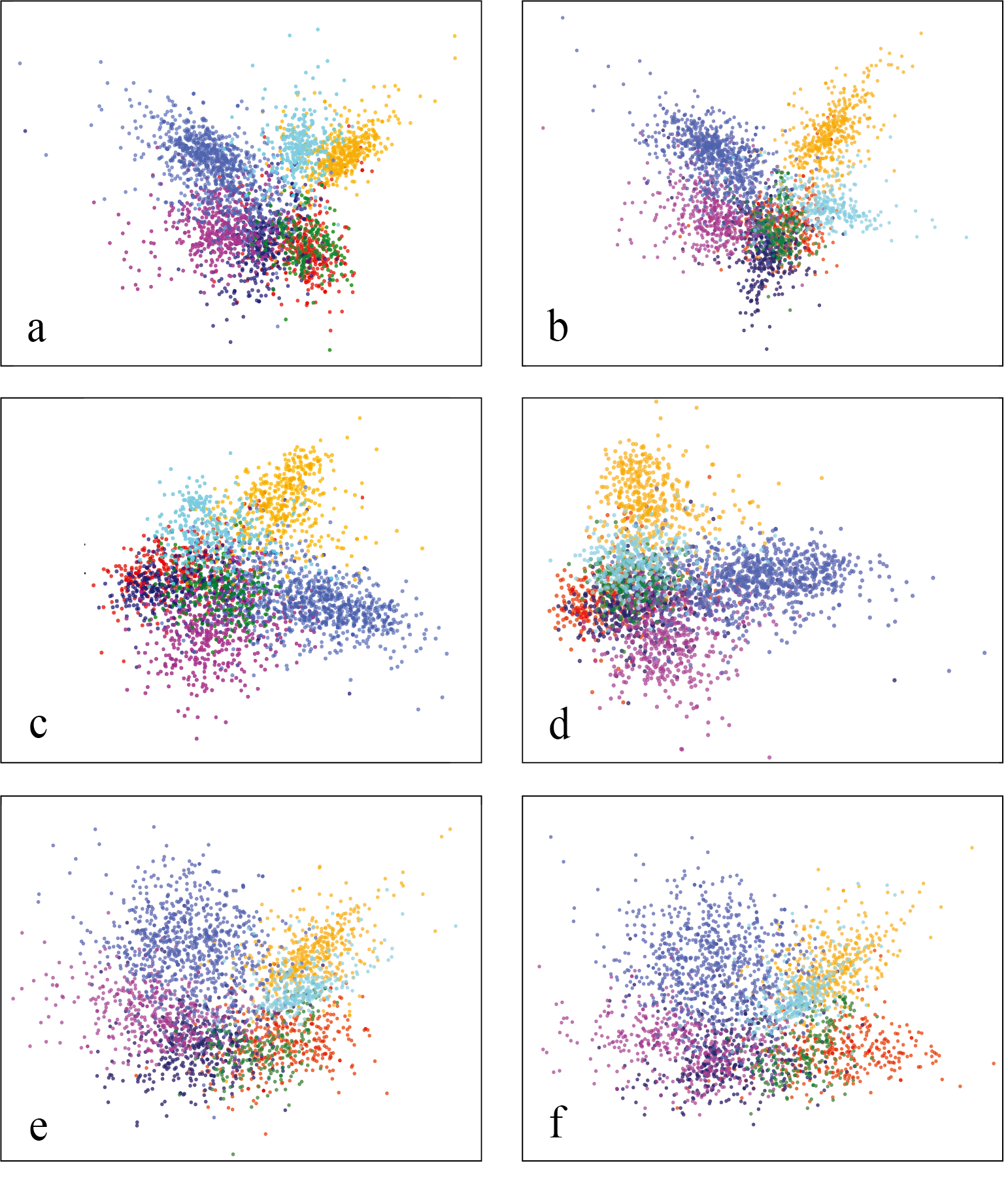}
      \caption{Visualizations for models' embedding layers. The model architectures corresponding to the markers in the figure are: (a) GCN, (b) GCN, (c) Chebnet, (d) GraphSage, (e) SRLIM, and (f) SRLIM.}
      \label{visual}
   \end{figure}

This section validates our theory about topology learning by visualizing some embedding layers of GNNs.
Figure 3 shows the PCA visualization of the embedding layers for six different models.
Some visualizations with additional mappings can destroy the integrity of the features.
Instead of doing additional mappings on the features, PCA selects the two most significant dimensions of the node features for visualization transformation.
Figures 3 (a) and (b) present the embedding layers of the same network architecture with different weights initialized for the model.
Figures 3 (a), (c) and (d) show the differences in embedding layers for models with different network architectures.
By comparison, we can see that the networks behave differently in their embedding layers when their network architectures are different.
When the networks have the same network architecture, the initialization of the weight parameters also leads to different node distributions, i.e., it does not maintain the topology of the input layer.

Figures 3(e) and (f) demonstrate the performance of SRLIM initialized with different weight parameters at the embedding layer.
It can be found that the node distribution of the trained SRLIM is relatively fixed in the embedding layer compared to the normal GCN.
In these two visualization figures, the nodes of the embedding layer find a balance between aggregation and maintaining the topology, so the distribution of nodes is divergent concerning those of the other models.
This proves that the topology of the input layer constrains the mapping process of SRLIM.
In addition, SRLIM can make the trained embedding layer converge to a stable form.
Thus, by visualization, we empirically demonstrate that general GNNs lose topological information during training, while SRLIM can retain topological information.

\section{Conclusion}
In this work, we focus on the transferability of attacks in untargeted gray-box attacks. 
We first discuss the impact of losing topological information on the reliability of the gradient provided by the surrogate model to the attacker.
Considering that the victim model is unknown, an attack with generalization should be based on the important prior of the topology of the nodes in the graph. 
We note that the model should learn the topology of the nodes in the input layer during the training of the surrogate model.
When topology information is propagated in the forward process, the gradient obtained by backpropagation will carry the topology in the input layer.
So far, we have summarized the need to preserve the topology for the surrogate model.
To this end, we propose SRLIM, a surrogate representation learning method using isometric mappings for preserving topology.
Considering that the graph data are distributed in non-Euclidean space, we introduce a similarity matrix based on the shortest path to estimate the topology of the nodes.
Subsequently, we use the isometric mapping to constrain the node similarity of the embedding layer to fit the node similarity of the input layer by Bergman divergence.

In the experimental section, we validate the effectiveness of our proposed method by experimenting with different settings of the untargeted poisoning gray-box attack. The results demonstrate that SRLIM gives the surrogate model the ability to learn the topology through isometric mapping, which improves the reliability of the gradient provided to the following attack model.

\bibliographystyle{ACM-Reference-Format}
\bibliography{egbib}

\end{document}